\documentclass[sigconf]{acmart}

\usepackage{booktabs} 
\usepackage{algorithmicx}
\usepackage{algorithm}
\usepackage{algpseudocode}

\copyrightyear{2024}
\acmYear{2024}
\setcopyright{acmlicensed}
\acmConference[SAC '24]{The 39th ACM/SIGAPP Symposium on Applied Computing}{April 8--12, 2024}{Avila, Spain}
\acmBooktitle{The 39th ACM/SIGAPP Symposium on Applied Computing (SAC '24), April 8--12, 2024, Avila, Spain}
\acmDOI{10.1145/3605098.3636000}
\acmISBN{979-8-4007-0243-3/24/04}

\acmArticle{4}
\acmPrice{15.00}
\newcommand{\name}{MultiConfederated Learning}  
\newcommand{\abbrname}{MCFL}  

\begin{document}
\title{\name: Inclusive Non-IID Data handling with Decentralized Federated Learning\\}


\author{Michael Duchesne}
\orcid{0009-0003-6453-508X}
\affiliation{
  \institution{École de Technologie Supérieure}
  \city{Montreal} 
  \state{Quebec} 
  \country{Canada}
}
\email{michael.duchesne.1@ens.etsmtl.ca}
\author{Kaiwen Zhang}
\orcid{0009-0003-6453-508X}
\affiliation{
  \institution{École de Technologie Supérieure}
  \city{Montreal} 
  \state{Quebec} 
  \country{Canada}
}
\email{kaiwen.zhang@etsmtl.ca}
\author{Chamseddine Talhi}
\orcid{0009-0003-6453-508X}
\affiliation{
  \institution{École de Technologie Supérieure}
  \city{Montreal} 
  \state{Quebec} 
  \country{Canada}
}
\email{chamseddine.talhi@etsmtl.ca}

\renewcommand{\shortauthors}{Duchesne et al.}

\begin{abstract}
Federated Learning (FL) has emerged as a prominent privacy-preserving technique for enabling use cases like confidential clinical machine learning. FL operates by aggregating models trained by remote devices which owns the data. Thus, FL enables the training of powerful global models using crowd-sourced data from a large number of learners, without compromising their privacy. However, the aggregating server is a single point of failure when generating the global model. Moreover, the performance of the model suffers when the data is not independent and identically distributed (non-IID data) on all remote devices. This leads to vastly different models being aggregated, which can reduce the performance by as much as 50\% in certain scenarios.

In this paper, we seek to address the aforementioned issues while retaining the benefits of FL. We propose \name: a decentralized FL framework which is designed to handle non-IID data.
Unlike traditional FL, \name\ will maintain multiple models in parallel (instead of a single global model) to help with convergence when the data is non-IID. With the help of transfer learning, learners can converge to fewer models. In order to increase adaptability, learners are allowed to choose which updates to aggregate from their peers. Finally, our solution still benefits from the global FL knowledge by leveraging various models trained by the network.
\end{abstract}

%
%



\maketitle

\section{Introduction}
\label{sec:intro}

Data privacy problems are growing everywhere. Therefore, it is critical to innovate with privacy-preserving machine learning techniques \cite{Liu_Ding_Shaham_Rahayu_Farokhi_Lin_2021,privacy_preserving}. Federated learning (FL) is an interesting solution because it enables collaborative training between data owners without requiring the centralization of data. FL works by training a model in parallel on data owners' devices \cite{fl_review_2019}. The data never leaves the device; instead, the trained model is sent to an orchestrator, which creates a global model by aggregating the received models. 

Although FL allows for the training of powerful global models using crowd-sourced data from a numerous learners without compromising their privacy, it faces several significant issues. Firstly, the presence of an orchestrator as a middleman introduces a single point of failure and requires trust from the clients, as aggregation is a critical component for producing an accurate global model. Moreover, the lack of guarantee regarding the distribution of data can diminish training effectiveness. This issue is common in FL setting since it is very likely that each learner will collect local data not representative of the entire set. Furthermore, due to privacy concerns, learners cannot share complementary data among themselves. This can decrease performance by up to 50\% in some cases depending on the level of skewness between learners \cite{wd-original-ref}. 

To address the issue of a central point of failure, Decentralised Learning (DML) systems like Swarm Learning \cite{sl_medical_imaging}, BFLC \cite{blockchain_dml} and Brain Torrent\cite{brain_torrent} have been proposed. These methods eschew the dependence on a central actor by connecting the learners using a peer-to-peer network or a blockchain. 

This paper aims to enhance the collaboration among siloed data owners by proposing a generalized decentralized approach of FL. Our solution considers the non-IID challenges in supervised and unsupervised settings, while preserving privacy-related benefits and ensuring participant autonomy.

We propose \name~(\abbrname), a decentralized FL approach. Contrary to traditional FL, various models exist simultaneously instead of having a single global model. Additionally, learners can select the most relevant models based on each model's performance using their local dataset. Each model acquires specialization by being trained only from a subset of all learners with data sharing similar features. Moreover, they can train multiple models and propose new updates by aggregating the updates of their peers. A forking mechanism is created as learners can choose different combinations of updates. This mechanism gives learners the necessary flexibility and independence to create a model that fits their data, while making their knowledge accessible to other peers. Moreover, we propose control mechanisms to converge to a single model when possible. The freedom given to every learner on the network to select models and updates to aggregate gives them the choice to select what is best for them while collaborating with other learners to achieve better results in less time. For this reason, we chose to use the term 'confederation' for our solution to reflect the analogy of power sharing among a group of sovereign entities collaborating to achieve a common objective.

The main contributions of this paper are the following:

\begin{enumerate}
    \item We propose a novel collaborative approach to decentralized Federated Learning, where multiple models are trained in parallel by different dynamic groups of trainers. Moreover, the network uses transfer learning between groups to facilitate convergence to a single model whenever possible. Models can be forked to satisfy every learner.
    \item We propose a selection algorithm to decide the set of models trainers should contribute to as well as an algorithm to select which updates should be aggregated to form the next models. 
    \item We compare the performance of our proposed approach with that of using the baseline FedAvg. Furthermore, we empirically demonstrate the benefits of the convergence feature of \name. 
\end{enumerate}

The structure of the paper is as follows. Section \ref{sec:rw} covers previous related papers. Then, Section \ref{sec:background} covers background concepts such as FL, DL, non-IID data, and multi-task learning. Section \ref{sec:proposed-solution} presents the architecture and details of our proposed solution \name. Section \ref{sec:results} shows the implementation details and the results of our evaluation in IID and non-IID scenarios. The paper concludes with Section \ref{sec:conclusion}.

\section{Related works} \label{sec:rw}
This section examines Federated Learning and non-IID data studies. While previous research addressed some introduced challenges, we will highlight the distinct advantages of our proposed approach over existing solutions.

\subsection{Non-IID data}

In \cite{wd-original-ref}, the author mitigates the impact of non-IID by sharing a public dataset every learner includes in their private training dataset. This does not respect our no sharing data policy. 

A Divide and Conquer FL algorithm was implemented in \cite{wd_dnd} with good results because their strategy was based on limiting weight divergence. We adopt a similar strategy aiming to limit weight divergence by grouping learners based on their preferred model. 

The authors in  \cite{overcoming_forgetting_in_fl} demonstrate that learning from a non-IID dataset is similar to learning multiple tasks. However, multi-task learning often suffer from forgetting when learning task sequentially. To overcome forgetting in FL, the authors used EWC \cite{ewc}, which modifies the loss function to mitigate the issue. \abbrname~ will fork the best model for a new task if it deteriorates the performance of the model for another task. This leads to model specialization. 

\begin{figure}[ht]
\centering
  \includegraphics[width=\linewidth]{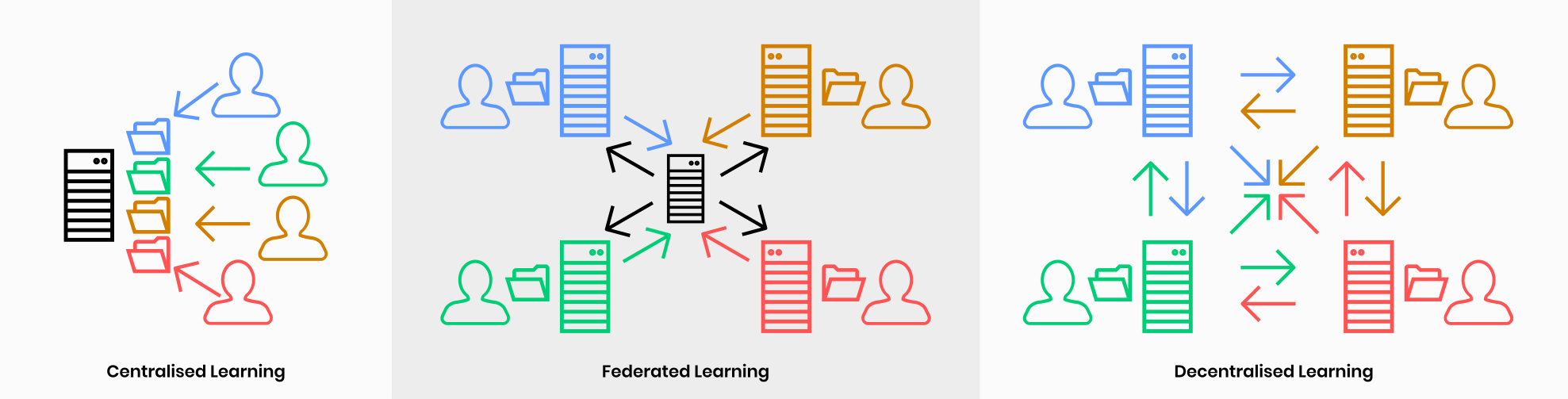}
  \caption{Overview of the three types of machine learning network deployments.}
  \label{fig:centralised-fed-swarm}
\end{figure}

\subsection{Federated Learning}

FL algorithms have been proposed, such as FedProx \cite{fedprox} and Scaffold \cite{scaffold}. The former attempts mitigates performance reductions from non-IID data by tolerating partial work and limiting the update range. The latter adds a bias during training to align learners to the same optimal parameters. This approach produces excellent results but it may not yield an optimal joint solution. These approaches are complementary to our proposed solution which could replace FedAvg.

Update selection and clustering have been tested and proven in the Federated Learning settings. HypClusters \cite{fl_hypclusters} show clustering can be beneficial for FL. In \cite{fl_clusters_k_means}, the orchestrator clusters clients according to their model using k-means algorithm. Hierarchical Clustering in \cite{fl_hierarchical_clustering} also clusters updates based on the Euclidian or Cosine distance method of the models. A direct acyclic graph (DAG) based approach was proposed in \cite{dfl_dag_specialization} that always merges two updates. Finally, \cite{fl_fmc}’s approach also builds a DAG with fork and merge abilities using EWC techniques. 

Our approach does not limit a learner to a single cluster and lets the learner switch between groups. For this reason, we also refer to clusters as groups in our paper. Nonetheless, most of the previous works do not consider weight divergence between the updates.

 Decentralized Learning is an adaptation of Federated Learning which does not use a central orchestrator (see Figure \ref{fig:centralised-fed-swarm}). Instead, this model connects all learners via a peer-to-peer (P2P) network as in \cite{brain_torrent} and delegates the aggregation of models to nodes within the networks. Blockhain has also been used in \cite{blockchain_dml} where a committee evaluates new iterations of the global model before appending them to the chain. Smart contracts were used in Swarm Learning \cite{sl_medical_imaging} to aggregate models. We opt for a P2P network. We also significantly change the training process as we train multiple variations of global models. It can be seen as if every node is running its own Federated Learning instance, so they keep total control of the training and aggregation process.

\section{Background} \label{sec:background}

In this section, we explore the essential techniques employed to implement the proposed solution successfully. We start by elaborating on Federated Learning and Decentralized Learning. Then, we detail the challenges of non-IID data and its relation to multi-task learning and transfer learning.

\subsection{Federated Learning}

As suggested by \cite{fl_google_keyboard}, Federated Learning, the central service does not have direct access to any data. It will randomly sample a subset of available devices and starts a training round with them. Then, selected devices have a defined amount of time to train the global model and send their update to the server. Once the round is over, the aggregate of the models becomes the new global model for the network. The server then sends the global model to the devices for the next round. The most straightforward aggregation is done using FedAvg. Assume that there are \textit{K} learners in round $n$ training the model $W_n$. The aggregator will determine a weight ($w^k$) for every model according to the number of samples on which it was trained and sum them all up. As a result, the sum becomes an average of all models (see Equation~\ref{eq:fedavg}).

\begin{equation}
W_{n+1}^g = \sum_{i=1}^{K} \frac{W_{n}^{k}}{w^k}
\label{eq:fedavg}
\end{equation}

With respect to communication costs, FL can be demanding depending on the size of the models. For example, in a network of $K$ clients with sampling rate $r$, the aggregator receives at most $K * r$ models. For this reason, the number of models transferred by the aggregator increases linearly with the number of clients. On the other hand, clients only need to send and receive at most one model per round.

\subsection{Decentralized Machine Learning}\label{background-dml}

Decentralized Machine Learning (DML) removes the centralized server for aggregation and relies on a Peer-to-Peer network or a distributed ledger. The removal of the aggregator pushes the aggregation responsibility to the network. Moreover, DML implementations like \cite{sl_medical_imaging} considers learners as sovereign actors as opposed to most FL implementations. Indeed, FL is often used to leverage data from clients by the aggregator's operator. 

As DML operates in a decentralised network, it tends to communicate more models compared to FL, because nodes need to send more models to coordinate and verify the aggregation has been done properly. Assuming the worst case scenario where every node is fully connected to all other nodes and participates to the training in a network of $K$ nodes, each of them needs to send and receive $K$ models every round. The lack of a centralized server implies that $K^2$ models are sent every round through the entire network since every node has to receive all models from its peers to perform the aggregation.

FL and DML can be executed in two variants: cross-devices and cross-silo. The former is where thousands of devices are expected to participate in training. The latter usually assumes that the clients are organizations, reducing the number of expected participants \cite{fl_cross_silo}. 

\subsection{Non-IID Data}
In both FL and DML scenarios, all learners rarely have the same distribution of samples, but must consider the difference in distributions because it significantly impacts model convergence. As shown by \cite{wd-original-ref}, weights from models trained on different distributions will tend to diverge more. While the models perform well individually, their aggregation results in a worse solution. Moreover, the divergence between models can lead to slower convergence of the global model. 

Data can be skewed in various ways; for example, the dataset may be unbalanced, so certain learners have more data samples from a label than others or classes completely absent \cite{non_iid_survey}. Another example is label preference skew, which refers to improperly labeled data or learners disagreeing on a label for similar samples.

Moreover, non-IID data is hard to distinguish from a covert poisoning attacks \cite{fl_fedcc}. Indeed, as the attacker can simply switch labels for a class, differentiating this from a learner with a preference skew is challenging. Additionally, Learners might not know the entire domain of the data which makes the process of detecting poinsonned models even more challenging. This uncertainty means that approaches ignoring updates like \cite{fl_krum} might exclude legitimate update from the network. Our proposed solution is more inclusive, since it allows learners to fork and train a separate model if some peers exclude them.

\begin{figure}[!ht]
  \includegraphics[width=0.9\linewidth]{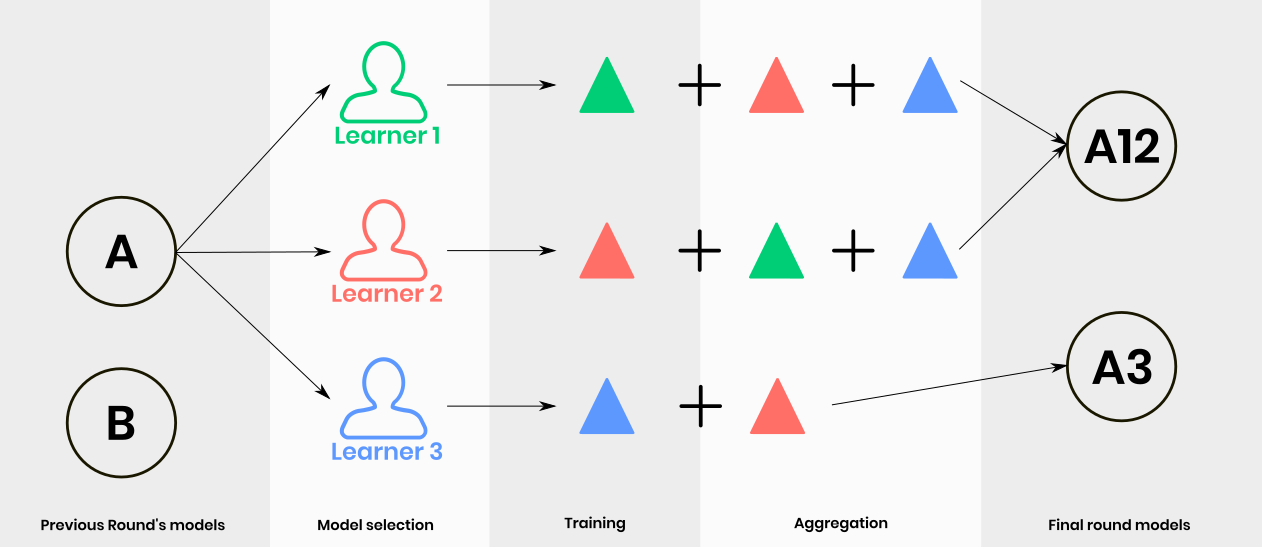}
  \centering
  \caption{Evolution of models during a single \name\ round.}
  \label{fig:dag-clusters-fork}
\end{figure}

\subsection{Multitask Learning}

In \cite{overcoming_forgetting_in_fl}, FL of a non-IID dataset is considered multitask learning. The different distributions are different tasks learned simultaneously and have commonalities which can be transferred\cite{transfer_learning} from one to the other. This ability has the effect of reducing training time and improving generalization which can manifest in a better performance for new tasks. 

When a NN attempts to learn tasks sequentially, it tends to forget previous tasks as it optimizes for the task currently learning~\cite{ewc}. Similarly, forgetting can also occur in FL when learners fine-tune the global model~\cite{fl_finetuning}. As a result, its performance increases for data in the local distribution, but decreases for the rest. Although improved performance on the learner dataset is desirable, one key advantage of FL and DML is creating a generalizable solution for all learners. 

\section{Proposed solutions} \label{sec:proposed-solution}
We propose \name, a decentralized FL approach using FedAvg to reduce the impact of non-IID data distributed among sovereign entities. It is a personalizable, rule-based and multifaceted approach using groups, transfer learning and weight divergence. In Section \ref{subsec:sys-model}, we present our assumptions and main architectural decisions. Then, we expand on the training process in \ref{subsec:net-init}. Section \ref{subsec:update-agg} and \ref{subsec:forks-man} elaborate on the forking mechanism during aggregation selection. Finally, we analyse communication costs in \ref{subsec:comms-cost}.

\subsection{System Model} \label{subsec:sys-model}

The proposed system model is a decentralized peer-to-peer approach based on Federated Learning. Each node acts as both the FL aggregator and the client, allowing them to train models and aggregate updates in synchronous round-based iterations. As a result, the system inherits the privacy properties of Federated Learning.

Since our approach is decentralized, learners cannot rely on a central server to assign groups and aggregate models. Therefore, learners use a DAG to track groups and models. We also introduce selection rules for nodes to autonomously choose the most appropriate groups.

Figure \ref{fig:dag-clusters-fork} presents a visual representation of the grouping model. In this figure, all learners decide to update model A, causing model B to die in this round. The model is then forked into two different models as learner 3 decides to exclude the update from learner 1.

The overview of a round can be divided into 3 steps:
\begin{enumerate}
    \item Learners select the most relevant models based on their own model performance metrics. The selection is explained in Subsection \ref{subsec:net-init}.
    \item Learners train the selected models and share their updated models. The training is also explained in Subsection \ref{subsec:net-init}.
    \item Learners aggregate updates and publish their selection of accepted updates. Subsection \ref{subsec:update-agg} explains the aggregation strategy in more detail.
\end{enumerate}

As discussed in \ref{background-dml} and evaluated in Section \ref{subsec:comms-cost}, due to communications increasing with the number of participants, we consider our approach to be aimed at cross-silo usage. Consequently, we expect nodes to be hosted on good hardware and always online with a reliable and fast connection. Moreover, nodes communicate through TCP, which guarantee the delivery of every message.

While our current solution is based on peer-to-peer communication, it could be implemented using a blockchain. Nevertheless, our main objective is to demonstrate the effectiveness and flexibility of the strategy, which does not require a blockchain.

Furthermore, our proposed approach prioritizes the autonomy of learners and provides them with considerable flexibility during the aggregation phase. Specifically, our approach allows them to selectively choose updates based on their weight divergence. We explain how this can create forks and how it can be handled in Subsection \ref{subsec:forks-man}. 

Since malicious actors and non-IID data cannot reliably be distinguished from covert poisoning attacks and we don't want to exclude learners falsely from the network, we do not consider them to be different from trainers with non-IID data. As we provide every learner with complete sovereignty regarding the final model published, they have the ability to exclude malicious actors when detected just like they can exclude divergent updates. This design decision is due to the difficulty of detecting malicious actors. Indeed, it is not a trivial task, but this approach lets honest nodes implement various and independent counter measures as selection rules which can be updated as improvements in this field are done. In case of false detection, the learner will still be publishing updates, but its models won't be included into aggregations.

\subsection{Network Initialisation and Learning Process}\label{subsec:net-init}

To begin with, the learners divide their dataset into three distinct subsets: training, validation, and test. The training and validation subsets are used during the training phase, whereas the test subset is used during the selection phase. This ensures the trainer will select generalized models. The splits should be performed randomly to maintain the same distribution, and in proportions that are considered acceptable by the learners. While a more robust mechanism to join the network could be used, learners simply broadcast a randomly generated identifier to a PubSub server. This is acceptable as we assume that no actors will attempt to impersonate other learners. Once every learner has connected, they begin the first round of training. During this initial phase, each learner trains a model on their training dataset and then shares this model with their peers as a new usable update. It is considered more privacy-friendly to share gradients, but we consider our solution applicable with any aggregation strategy.

At the beginning of every subsequent round, the learners enter the selection phase. In this phase, learners establish relevant metrics for their use case. For example, these could be accuracy, loss, F1, etc. During this phase, each learner evaluates all models on their test dataset and records the corresponding performance metrics. Based on these metrics, each learner selects the best-performing models to train during the round. The selection of models is left to the discretion of each learner, who can use any relevant metrics to make their decisions. As the learning process is stochastic, some models will perform better than others on data that they have never seen before. Consequently, some learners may switch to a model in which none of their updates was aggregated, thereby consolidating learners around the best models and enhancing generalization capability.

Learners can train more than one model to transfer knowledge from their data to other models through transfer learning. As previously discussed, learning from a non-IID distribution can be considered as a multitask problem, and transfer learning can be applied to help improve the model's performance across tasks. The solution's implementation for transfer learning involves training the models of other groups as the learner would do for its own group. While more sophisticated transfer learning techniques could be used, our approach demonstrates this feature's relevance within the system. After training, the learner shares the updates of all trained model. Finally, during the aggregation phase, learners select the updates they want to aggregate based on their aggregation rules. Subsequently, the learners communicate their choice with the rest of the network. Once the learners have received the choices from their peers, the round is finished, and they start another round in the selection phase. 

\subsection{Updates Aggregation} \label{subsec:update-agg}

The proposed approach sacrifices the usual generalization of a global model by training multiple models to address the non-IID issue. The goal is to group learners around models rather than around data. As a result, the non-iidness of data within the group should be reduced. By reducing non-iidness, we reduce update divergence and accelerate convergence. In other words, learners should group to train models where their updates are least divergent. Furthermore, as groups have similar data, they are more likely to find a common optimal solution. Ultimately, each group develops a model tailored to its members.

\begin{figure*}
  \includegraphics[width=0.9\linewidth]{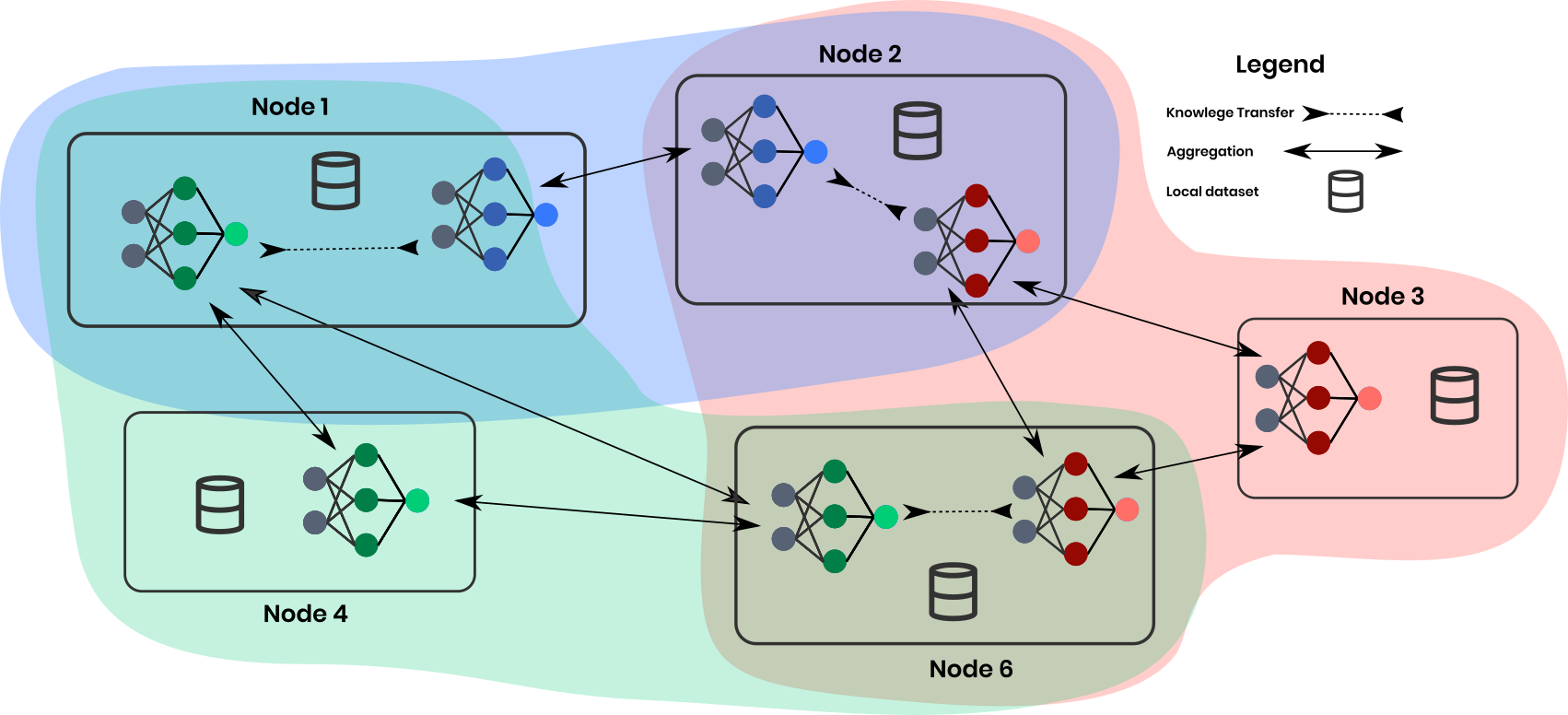}
  \centering
  \caption{An overview of a \name\ network comprising 5 nodes, which are divided into 3 distinct groups.}
  \label{fig:network-architecture}
\end{figure*}

Unlike previous clustered networks, our approach does not confine learners to a single cluster \cite{fl_clusters_k_means,fl_fmc,fl_hypclusters}. Learners have access to the updates and models of every group in the network. Moreover, they can contribute to multiple models simultaneously in an attempt to transfer knowledge between the groups and gradually expose other models to new data. Figure \ref{fig:network-architecture} depicts the network architecture, showing how learners group for aggregation and transfer of knowledge between groups. In this hypothetical network, nodes are split into three different groups, with some nodes belonging to two groups simultaneously. This allows those nodes to engage in transfer learning, to potentially converge to a single group.

The network forms a DAG representing the evolution of the models by training a parent model, which corresponds to a node in the graph. Learners always submit their updates to a parent model. Furthermore, learners can select a combination of updates to be aggregated for the next round. It is important to note that each learner selects updates and publishes their selection. Indeed, they have the freedom to choose any updates based on their criteria and assessment of reliable actors.

Each learner aggregates the updates using FedAvg. The proposed approach uses the FedAvg algorithm for its simplicity and its deterministic nature. Indeed, with access to all models and updates, learners can independently perform aggregations for each model version proposed by peers, ensuring uniformity in the resulting models. This ability spares the communication of additional models. 

Flexibility in aggregation allows to creates different models, as different learners choose different combinations of updates. The combinations create a forking mechanism where a parent model can have many children. As every learner can fork a parent model, it guarantees they will have at least one satisfactory model for the round as they can choose their own version from the previous round. 

\subsection{Forks Management}\label{subsec:forks-man}

As mentioned earlier, the proposed approach gives a lot of freedom to the learners to select which models they train and which updates they decide to aggregate. The combinations created lead to more forks, increasing training time, and reducing training efficiency as learners spend more time testing models.

The network aims to balance the performance of models, the quantity of models, and their generalization capabilities. The proposed approach reaches this balance with a model selection rule that encourages the selection of more popular models.Specifically, the cluster selection algorithm computes a score based on a test metric and the number of updates (defined as popularity) each model received in previous rounds. Assuming there are \textit{n} models, the algorithm selects the $\sqrt{n}$ top-performing models for training. We employ a square root to ensure the number of trained models does not scale linearly with the number of models since selecting more models increases the duration of the round. On the other hand, it improves transfer learning between groups to select more models. An overview is shown in Algorithm \ref{alg:cluster-selection}.

\begin{algorithm}
    \caption{Models Selection}\label{alg:cluster-selection}
    \begin{algorithmic}[1]
        \Procedure{selectBestModels}{models}
            \State $modelResults \gets testModels(models) $
            \For{\textit{i} \textbf{in} \text{range(}\textit{models.length}\text{)}}
                \State $popScale \gets \text{sqrt(popularity(} \textit{models[i]} \text{))} $
                \State $modelScores[i] \gets \textit{modelResults[i]} \text{ * } popScale $
            \EndFor
            \State $rankedModels \gets models.orderBy(modelScores)$
            \State $nModels \gets sqrt(models.length)$
            \State $\textbf{return } rankedModels[0: nModels]$
        \EndProcedure
    \end{algorithmic}
\end{algorithm}

After completing local training and sharing their updates, learners must select updates for aggregation, using aggregation rules to exclude potentially detrimental ones. As discussed earlier, weight divergence is a key factor in the decreased performance of aggregated models, particularly in scenarios where data is non-uniformly distributed. To address this issue, we have devised an aggregation rule focusing on weight divergence, which is calculated using Equation \ref{eq:wd} from \cite{wd-original-ref}. While it is possible to develop additional rules to meet specific learner needs (e.g., a rule for Byzantine resistance based on the trustworthiness of other learners), our focus is on accelerating model convergence in non-IID settings by mitigating weight divergence.

\begin{equation}
Weight Divergence = \frac{ || {W}^{PeerUpdate} - {W}_{}^{LocalUpdate} ||}{{W}_{}^{LocalUpdate} }
\label{eq:wd}
\end{equation}

As shown in Algorithm \ref{alg:aggregation-selection}, the rule requires each node to pick a configurable value called the tolerance. This value, representing a number of standard deviation, will impact the divergence between updates tolerated by the trainer. Setting it between 2 and 3 typically includes 95\% to 99\% of updates. This serves as dynamic threshold to filter out updates which are too divergent. A lower tolerance will lead to less divergent updates being aggregated, which will make the model more performant without any fine-tuning. The algorithm first calculates the weight divergence between the learner's update and each received update, and then computes the median and standard deviation of the weight divergences. Using these statistics and the tolerance value, it selects the updates that are within a range of the learner's update. The goal is to avoid aggregating updates which are too divergent. A higher tolerance includes more updates and reduce forks, but will lead to the aggregation of more divergent models. On the other hand, a lower tolerance level excludes more updates from the aggregate, increasing the potential combinations which results in more forking.

\begin{algorithm}
    \caption{Update Selection}\label{alg:aggregation-selection}
    \begin{algorithmic}[1]
        \Procedure{selectUpdates}{myUpdate, updates, tolerance}
            \For{\textit{i} \textbf{in} \text{range(}\textit{updates.length}\text{)}}
                \State $divergences[i] \gets wd(myUpdate, updates[i]) $
            \EndFor
            \State $med \gets median(divergences)$
            \State $maxDiv \gets med + std(divergences) * tolerance$
            \State $selIndex \gets 0$
            \State $selected[selIndex] \gets myUpdate$
            \For{\textit{i } \textbf{in } \text{range(}\textit{updates.length}\text{)}}
                \If{divergences[i] \textbf{ less then } maxDiv}
                    \State $selIndex \gets selIndex + 1 $
                    \State $selected[selIndex] \gets updates[i] $
                \EndIf
            \EndFor
            \State $\textbf{return } selected$
        \EndProcedure
    \end{algorithmic}
\end{algorithm}

\subsection{Communication Costs} \label{subsec:comms-cost}

In \abbrname, the nodes communicate models with each other through a peer-to-peer network. As a result of training multiple models, the learners have to transfer more models compared to the traditional DML network for each round. For example, in a network of $K$ nodes each training $M$ models, a single node will send $M * K$ models. Although it increases communication cost linearly with the number of models being trained, we do not consider it to be more problematic than for DML, because the time for a round in \name\ also increases linearly with the number of models trained. As both round time and communication increases proportionally, model transfer can be done on a longer timeframe without altering the rate.

In addition, learners have the overhead of communicating their aggregate models when compared to FedAvg. However, they only need to share their selection, not models. Indeed, because the aggregation is deterministic, a trainer can replicate the aggregated model by aggregating the update selection from their peers. For this reason, we consider the communication overhead as negligible when compared to traditional DML.

\section{Results} \label{sec:results}
This section assesses the performance of the proposed approach. First, the experimental setup is described. Second, the proposed approach is compared to the baseline solution. Finally, the results show that multiple models can give better performance and faster convergence.

\subsection{Experimental Setup}

We implemented the system in Python 3.9 using Pytorch Lightning 1.8, a framework built on Pytorch. The learners have access to an MLFlow instance to store information about their training. The MLFlow instance relied on a PostgreSQL database to store its data. All nodes were deployed in Docker containers on a bare metal server running Ubuntu 22.04. The server uses dual Intel Xeon Gold 5118 CPU totalling 48 cores with 126GB of RAM. All the nodes in the network are doing training and aggregation. Nodes send messages peer-to-peer through an MQTT instance configured to broadcast messages. Moreover, learners do not communicate their models directly. Instead, they save their models to a location accessible to other peers and the model's url is shared.  

\textbf{Dataset:} The network was tested using two computer vision problems with MNIST and FashionMNIST datasets. MNIST consists of handwritten digits while FMNIST contains pictures of clothing items. While these datasets are typically balanced, we intentionally distributed samples in various non-IID ways. The specifics of these distributions are detailed prior to the experimental results. This methodology for generating non-IID data has been previously used in many papers such as \cite{fedprox} and \cite{fed_nova}. These dataset are ideal for concept validation using the same input dimensions and well-known comparable benchmarks. At the start of every experience, the dataset is split among all the learners. Data samples are not reused between learners. All the tests are run using 38 learners for 102 rounds with two epochs per round.

\textbf{Models:} The LeNet\cite{lenet} is used for the classification tasks. While the LeNet model works for supervised problems, we also want to demonstrate the ability of the network to work with unsupervised learning problems. For this reason, we chose to train autoencoders. They are a type of neural network that tries to replicate its inputs to the outputs by encoding the features in smaller latent spaces. The architecture of our autoencoders repurposes the first two convolutional layers of the LeNet model as the encoder and mirrors them for the decoder. The baseline local training achieved a satisfying level of accuracy with ten epochs.

\begin{table}[htbp]
\caption{Performance of local learners with all samples}
\begin{center}
\begin{tabular}{|c|c|c|c|c|}
\hline
            & \multicolumn{2}{c|}{MNIST} & \multicolumn{2}{c|}{FMNIST}  \\
\hline
Model       & Accuracy & Loss           & Accuracy & Loss             \\
\hline
LeNet       & 97\%     & 0.089          & 83\%     & 0.46             \\
\hline
LeNet CAE   & -        & 0.006          & -        & 0.015      \\
\hline
\end{tabular}
\label{tab:local-performances}
\end{center}
\end{table}

\textbf{Baseline Solution:} We implemented the baseline solution by reusing the architecture of our proposed approach. As a result, the baseline solution is a decentralised learning network in which all learners participate in each round, performing aggregation with the FedAvg algorithm.

\textbf{Metrics:} The performance of the models is measured on the learners dataset after the aggregation phase, without fine-tuning. For classification problems, we use accuracy as shown in Equation \ref{eq:acc}. TP stands for true positive, TN stands for true negative, FP stands for false positive, and FN stands for false negative.

\begin{equation}
 Accuracy=\frac{TP + TN} {TP + TN + FP + FN}
\label{eq:acc}
\end{equation}

The Mean Squared Error (Equation \ref{eq:mse}) is used for the autoencoder task. $y_i$ stands for the target prediction while $\hat{y}_i$ stands for the model output.

\begin{equation}
Mean Squared Error =  \frac{1} {N} \sum_{i=1}^{n}(y_i-\hat{y_i})^2
\label{eq:mse}
\end{equation}

\subsection{IID Experiments}
The first results in Table \ref{tab:iid-networks-performance} show the weight divergence selection mechanism is helpful to outperform FedAvg even on IID datasets. For this experience, the dataset was distributed uniformly. \name\ outperformed the FedAvg implementation, with the worst performer from \name\ almost outperforming the average learner in the baseline solution. As a result of the limited weight divergence between aggregated updates, the aggregated models’ performances are more predictable, and the gap between the worst and best results is smaller than with FedAvg. Moreover, the selection algorithm only selects the best-aggregated cluster, which helps with a slightly faster convergence rate in this scenario. In fact, the average performance is 20\% higher with the proposed approach for the first round but fades to 2-3\% improvements by the fifth round.

\begin{table}[htbp]
\caption{Performance Networks on IID Dataset}
\begin{center}
\begin{tabular}{|c|c|c|c|c|}
\hline
Algorithm       & Minimum       & Average       & Maximum       \\
\hline
FedAvg          & 96.2\%        & 98.2\%        & 99.5\%        \\
\hline
\abbrname        & 98.1\%        & 99.2\%        & 99.8\%        \\
\hline
\end{tabular}
\label{tab:iid-networks-performance}
\end{center}
\end{table}

\subsection{Non-IID Experiments}
\begin{figure}
  \includegraphics[width=\linewidth, trim=0 3cm 0 0]{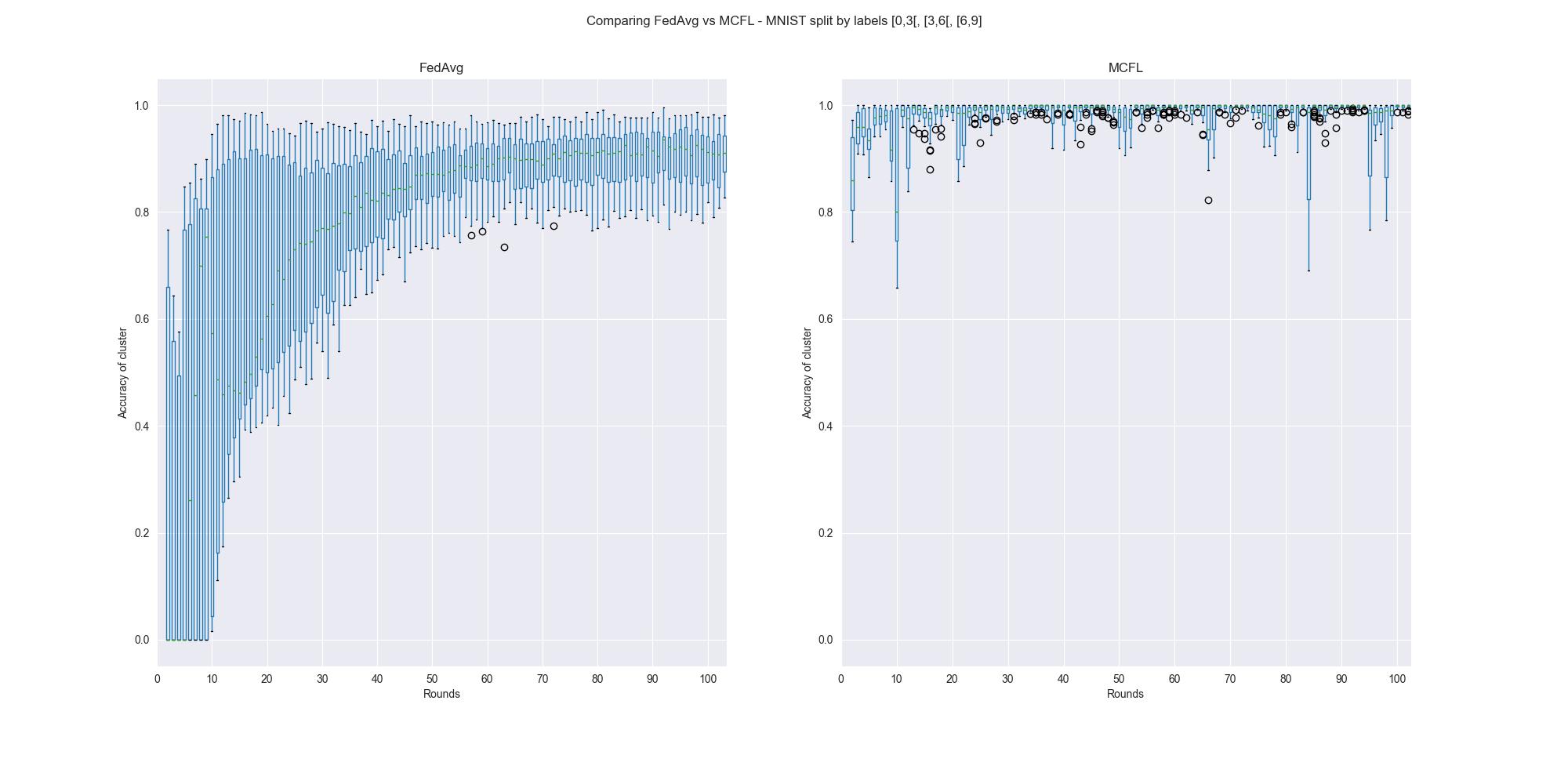}
  \caption{Accuracy of learners using FedAvg vs. \name. }
  \label{fig:non-iid-networks-performances}
\end{figure}

The proposed approach was also tested on various non-IID cases. In our first non-IID experiment, we split the learners into three subsets and assign classes exclusively to one of the subsets. The subsets of learners split data equally amongst them. As a result, each subset becomes IID sharing samples from the same class, while the entire dataset is distributed in a non-IID way through the network. Figure \ref{fig:non-iid-networks-performances} shows the accuracy of the aggregated model on the learner’s dataset. It is a challenging scenario for the baseline solution to converge. Even after 100 rounds, the global model is not performing as well as in an IID scenario, with the worst learners having 82\% accuracy. 

On the other hand, the learners from the proposed approach end up grouping properly and contributing to models which perform well for their data. They manage to accurately label data from their dataset with higher accuracy than the IID scenario, as classifying the subset is an easier task with fewer labels than the baseline task. However, their accuracy falls if tested on a balanced dataset as the models cannot label data outside of the group’s distribution. This case fails to do transfer learning as the groups have very different distributions. 

We empirically tested the effects of various tolerance thresholds explained in \ref{subsec:update-agg} on a new non-IID scenario. This time, we distribute the labels to the learners using a different normal distribution for every learner. This case does not have distinct groups; instead, each learner has overlapping data with many other learners.

\begin{figure}
  \includegraphics[width=\linewidth]{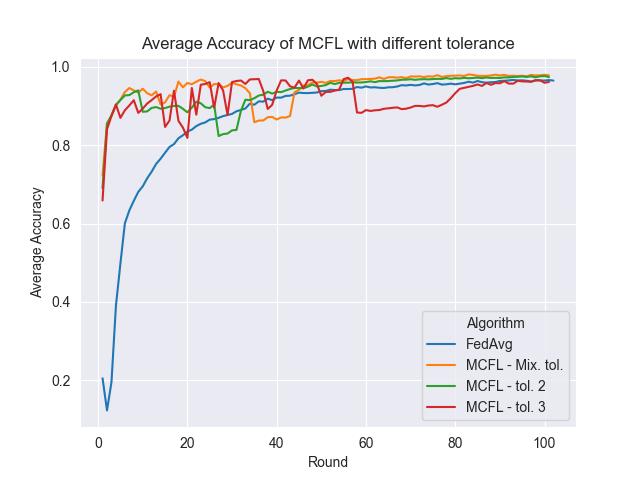}
  \caption{Average accuracy of multiple update selection tolerances with data distributed with no clear groups}
  \label{fig:tolerance-analysis}
\end{figure}

Figure \ref{fig:tolerance-analysis} shows the accuracy of the network for different tolerance thresholds. Many spikes can be seen in the figure which are the results of the popularity modifier to encourage merging and transfer of knowledge between models. Those degradations only impact a single cluster and it quickly recovers. The various combinations of updates mimic the process of evolution and natural selection. Indeed, each learner proposes a new iteration of the model which is in most is case better than the previous, but some times it can be worst. When it happens, it will not get selected for the next round and will go extinct. In this particular experiment, it was due to the popularity modifier which encouraged two groups to merge. Although this merger did not improve performance, the benefit of cluster convergence, leading to more generalized models, is considered to outweigh the temporary performance degradation.

\begin{figure}
  \includegraphics[width=\linewidth]{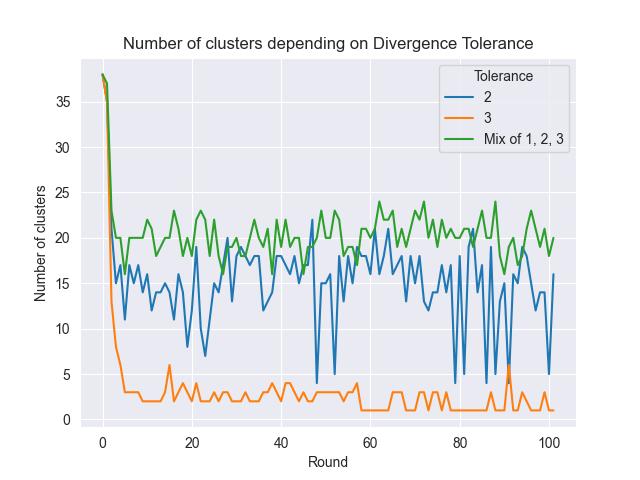}
  \caption{Number of models generated using various update selection tolerances with data distributed with no clear subsets}
  \label{fig:cluster_count}
\end{figure}

A higher configured tolerance threshold increases the aggregated model performance volatility as it includes more divergent updates. The mix of threshold demonstrates interesting performance characteristics, exhibiting less volatility in both accuracy and cluster counts in Figure \ref{fig:cluster_count}.

Moreover, Figure \ref{fig:cluster_count} shows a lower tolerance threshold results in increased forking. The figure also reveals that a tolerance of 3 makes the network converge to a single cluster, unlike other scenarios. This result makes sense as three standard deviations mean learners will include most updates. In a real world scenario, we do not expect an entire network to pick the same rules and tolerance threshold. This experiment only demonstrates how groups behave in different scenarios. 

The grouping approach is also effective with non-supervised problems as we demonstrate by training autoencoders. The experiment splits the learners into two groups; the first group is assigned MNIST, the second FMNIST, and both datasets are divided between all their learners. By training the LeNet CAE model with an aggregation tolerance of 3 standard deviations, the network successfully grouped to train two models with different performances for the two datasets. The two groups outperformed the reference global model by 24.5\% for MNIST and 17.5\% for FMNIST. 

This experiment demonstrates that Mean Squared Error (MSE) can also be effective in selecting the most suitable model for the learners' data, further confirming the flexibility of this approach.

\subsection{Results Discussion}
The results show that it is possible to have models converging much faster by grouping learners. We also demonstrated the ability of the network to converge to a single model. It is hard to make a fair comparison as MCFL trains a variable amount of models per round and does it in a parallel way. The experiments completed are also pretty restricted and quite extreme cases. Although it does prove the solution can achieve what it aims to do, it may not outperform FedAvg as much in other cases.

We believe the solution is interesting as it offers a few advantages and proposes a very flexible framework paving the way for a more robust decentralized learning future. One of the advantages standing out is the ability for every trainer to specialize a forked model and still contribute with meaningful updates which will always be included into the network. Moreover, it gives the liberty to implement various rules to choose their models and the accepted updates. Most importantly, it removes the trust requirements for the aggregator as the trainer will do its own aggregation.

There are additional approaches which could take into account the computational limitations of devices and bandwidth constraints. This methodology allows for versatility in the selection of groups by trainers. Consequently, trainers could employ specific heuristics, which might not yield optimal results, but would offer a balance between the time required per training round and overall model performance.

\section{Conclusions}
\label{sec:conclusion}
This paper introduces \name, a decentralized Federated Learning solution without a central aggregator, specifically designed to handle non-IID data. The groups formed during training were able to train a model that fit their data quickly. With more time and if advantageous, they can converge toward a single model. The key elements to achieve this are training multiple models, transfer-learning, well-thought selection rules and limiting weight divergence between updates. The results showed that the proposed approach worked effectively in two challenging non-IID scenarios. 

For future work, we wish to explore how to train models to generate synthetic datasets to train new models locally. Additionally, we are interested in using synthetic datasets and their generators to perform knowledge distillation to create a global model locally. Finally, we think that our work sets a good path forward for a robust byzantine-resistant decentralized machine learning network.

\section{Acknowledgement}
This work was supported in part by funding from the Innovation for Defence Excellence and Security (IDEaS) program from the Department of National Defence (DND)

\bibliographystyle{ACM-Reference-Format}
\bibliography{sample-sigconf} 

\end{document}